\documentclass[runningheads]{llncs}

\usepackage{eccv}

\usepackage{soul}

\usepackage{eccvabbrv}

\usepackage{graphicx}
\usepackage{booktabs}

\usepackage[accsupp]{axessibility}  

\usepackage{hyperref}

\usepackage{orcidlink}

\usepackage{amsmath,amsfonts,bm}

\def\eqref#1{equation~\ref{#1}}

\def\1{\bm{1}}

\DeclareMathAlphabet{\mathsfit}{\encodingdefault}{\sfdefault}{m}{sl}
\SetMathAlphabet{\mathsfit}{bold}{\encodingdefault}{\sfdefault}{bx}{n}

\usepackage{wrapfig}
\usepackage{graphicx}
\usepackage{booktabs}

\usepackage{lipsum}

\newcommand\blfootnote[1]{%
  \begingroup
  \renewcommand\thefootnote{}\footnote{#1}%
  \addtocounter{footnote}{-1}%
  \endgroup
}

\newcommand{\expect}{\mathop{\mathbb E}}%
\definecolor{PZH_color}{RGB}{0, 102, 204}

\newcommand{\bs}{\mathbf s}
\newcommand{\ba}{\mathbf a}
\newcommand{\bo}{\mathbf o}
\newcommand{\cS}{\mathcal S}
\newcommand{\cA}{\mathcal A}

\newcommand{\cI}{\mathcal I}

\newcommand{\bshistory}{\bs_{\text{history}}}
\newcommand{\bspredict}{\bs_{\text{pred}}}
\newcommand{\bsGT}{\bs_{\text{GT}}}
\newcommand{\Tpred}{T_{\text{pred}}}

\newcommand{\citep}[1]{\cite{#1}}
\newcommand{\citet}[1]{\cite{#1}}

\usepackage[noend,linesnumbered]{algorithm2e}
\let\oldnl\nl
\newcommand{\nonl}{\renewcommand{\nl}{\let\nl\oldnl}}%

\SetCommentSty{mycommfont}
\SetKwComment{Comment}{$\triangleright$\ }{}

\usepackage{hyperref}
\usepackage{url}

\begin{document}

\title{
Improving Agent Behaviors with RL Fine-tuning for Autonomous Driving
}

\titlerunning{
Improving Agent Behaviors with RL Fine-tuning for Autonomous Driving
}

\author{Zhenghao Peng\inst{1}\textsuperscript{*}
\and
Wenjie Luo\inst{2}
\and
Yiren Lu \inst{2}
\and
Tianyi Shen \inst{2}
\and
Cole Gulino \inst{2}
\and
\\
Ari Seff \inst{2}
\and
Justin Fu  \inst{2}
}

\authorrunning{Z.~Peng et al.}

\institute{
\textsuperscript{1}UCLA,
\textsuperscript{2}Waymo
}



\maketitle

\begin{abstract}
A major challenge in autonomous vehicle research is modeling agent behaviors, which has critical applications including constructing realistic and reliable simulations for off-board evaluation and forecasting traffic agents motion for onboard planning. 
While supervised learning has shown success in modeling agents across various domains, these models can suffer from distribution shift when deployed at test-time. In this work, we improve the reliability of agent behaviors by closed-loop fine-tuning of behavior models with reinforcement learning. Our method demonstrates improved overall performance, as well as improved targeted metrics such as collision rate, on the Waymo Open Sim Agents challenge. Additionally, we present a novel policy evaluation benchmark to directly assess the ability of simulated agents to measure the quality of autonomous vehicle planners and demonstrate the effectiveness of our approach on this new benchmark.
  \blfootnote{
  \textsuperscript{*}Work done as an intern at Waymo.}
\end{abstract}






\begin{wrapfigure}{r}{0.55\textwidth}
\centering
\vspace{-3em}
\includegraphics[width=\linewidth]{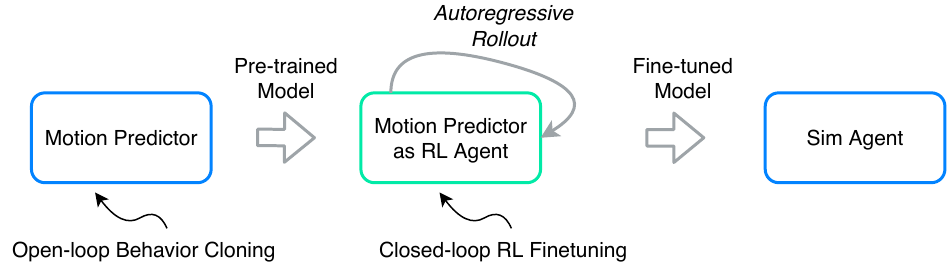}
\caption{We propose to fine-tune a pre-trained motion prediction model with closed-loop reinforcement learning.}
\label{fig:diagram}
\vspace{-1em}
\end{wrapfigure}

\section{Introduction}

Transformer-based architectures have demonstrated state-of-the-art performance in a variety of tasks in language~\citep{openai2023gpt4}, vision~\citep{ramesh2022hierarchical}, and robotics~\citep{zitkovich2023rt}.
The success of these models is credited to a widely adopted ``pre-training and fine-tuning'' scheme~\citep{zhou2023comprehensive}. In the pre-training phase, the model acquires knowledge from a very large amount of training data; during fine-tuning, the model behaviors are rectified to align with human preferences and expectations.
While supervised learning can be used for fine-tuning, previous work has shown superior performance with reinforcement learning (RL) fine-tuning in language tasks~\citep{ouyang2022training} and text-to-image generation~\citep{black2023training}. In autonomous driving (AD), given the abundance of human driving data, a natural question arises: Can we leverage the popular ``pre-training and RL fine-tuning'' strategy to effectively model agent behaviors?

In this paper, we investigate the viability of applying the ``pre-training and RL fine-tuning'' paradigm to model the behaviors of traffic agents in AD scenarios. Such models can be applied in critical AD tasks such as simulation agents (sim agents)~\citep{montali2023wosac}, enabling high-fidelity simulation systems for off-board evaluation, and behavior prediction for surrounding traffic participants, facilitating onboard planning.

\begin{figure}[!t]
\includegraphics[width=\linewidth]{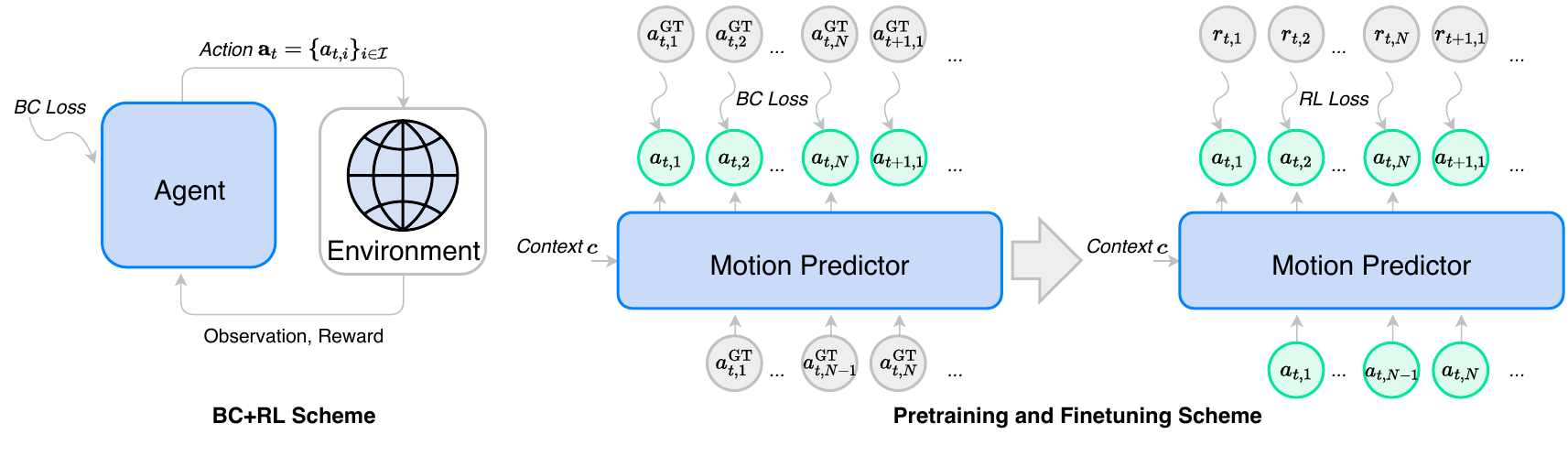}
\caption{
\textbf{Left}: The agent is trained from scratch using a combined Behavioral Cloning (BC) and Reinforcement Learning (RL) approach. Without pre-training on large datasets, the agent must simultaneously explore the environment and develop its capabilities from scratch.
\textbf{Right}: The agent undergoes a two-phase training scheme. Agent acquires a foundational skill set from aligning its actions (green) with ground truth data in pre-training (gray). The fine-tuning through RL refines the agent behaviors in the autoregressive rollout.
}
\label{fig:diagram2}
\end{figure}

Behavioral cloning (BC), or training an imitative model using supervised learning on demonstrations~\citep{argall2009survey}, has been the predominant approach for learning driving agents~\citep{codevilla2018end,wang2023multiverse}.
While BC provides supervision for modeling realistic behavior, during closed-loop simulation, the agent behaviors can deviate from the training distribution, known as the ``covariate shift'' issue~\citep{ross2011reduction}.
Moreover, BC lacks the ability to explicitly incorporate human preferences, expectations and  constraints.
For example, safety-critical events, such as collisions, are only implicitly discouraged in the BC loss due to their rarity in human driving datasets. In situations where a collision is likely to occur, the model may only have limitated examples to learn from.  
RL fine-tuning can address these limitations.
Firstly, RL learns from closed-loop synthetic rollouts, addressing the covariate shift problem as the reward function penalizes actions leading to future trajectories that diverge from ground-truth.
Secondly, explicit objectives can be incorporated into the reward function so the agents can learn to align with human preferences and expectations. 

Inspired by the success of fine-tuning large language models to align with human preferences, we apply the ``pre-training and RL fine-tuning'' scheme to training behavior models for sim agents.
As demonstrated in Fig.~\ref{fig:diagram}, we can fine-tune a pre-trained model via a simple on-policy RL approach with autoregressive rollouts.
We propose a simple reward function that not only enables the model to satisfy human preferences on the agent behaviors, but also maintains human likeness.

Our experiments on the Waymo Open Sim Agent Challenge (WOSAC)~\citep{montali2023wosac} demonstrate that RL fine-tuning significantly improves the reliability of the agent behaviors, especially 
in terms of collision avoidance.
An important application of the learned behaviors is in actuating the traffic agents in AD simulation. We study the reliability of the learned models in a novel planner evaluation benchmark.
The intuition is that a simulator with a more realistic sim agents model should provide more reliable evaluation across ego AD planners. 
With this insight, we use different sim agent models to control the traffic agents and assess the performance of a predefined set of AD planners. By comparing the planners' estimated performance as evaluated by the sim agents model against their known performance ranking, we find that our fine-tuned models provide more accurate planner evaluations. This indicates that our approach is beneficial for testing autonomous driving planners.
The main contributions of this work are:

1) We propose to apply the popular ``pre-training and RL fine-tuning'' paradigm commonly used for large language models (LLMs) to the autonomous driving behavior modeling problem, demonstrating the effectiveness of closed-loop fine-tuning a Transformer-based architecture on the Waymo Open Motion Dataset (WOMD)~\citep{waymo_open_dataset}.

2) We demonstrate that an on-policy RL algorithm with a simple reward function can successfully preserve the realism in the dataset while aligning human preferences on safety and reliability.

3) To better evaluate the performance of sim agents models, we propose a novel planner evaluation task and demonstrate that our method can significantly improve the performance of the sim agents models in terms of its capability to assess the AD planners.

\section{Related Work}

\subsection{Pre-training and Fine-tuning of Transformer-based Models}
Transformer-based models
have been applied to various domains such as text generation~\citep{brown2020language}, image generation~\citep{ramesh2022hierarchical}, robotics~\citep{zitkovich2023rt}, drug discovery~\citep{mendez2022mole}, disease diagnose~\citep{zhou2023foundation}, and generalist medical AI~\citep{moor2023foundation}.
Many large Transformer-based models are trained in the ``pre-training then fine-tuning'' manner, where supervised fine-tuning~\cite{zhou2023comprehensive} or reinforcement learning with human feedback~\cite{ouyang2022training} holds the promise to align the model behaviors to human preferences. 
In the autonomous driving domain, similar 
Transformer-based architectures have been applied to various tasks, ranging from perception~\citep{min2023uniworld}, motion prediction~\citep{nayakanti2023wayformer}, self-driving policies~\citep{hu2023planning} and simulation~\citep{suo2021trafficsim,zhang2023trafficbots,zhong2023guided}.
In this work, we focus on the motion prediction problem, where predictors forecast the future trajectories of the target agents by observing history information~\citep{nayakanti2023wayformer,shi2022motion,shi2023mtrpp}.
Unlike foundational models in other domains such as large language models~\citep{openai2023gpt4} and vision language models~\citep{lu2019vilbert},
motion prediction models are most commonly trained via supervised learning and rarely fine-tuned to better boost alignment with human preferences.

\subsection{Behavior Modeling for Autonomous Driving}
Modeling the behavior of traffic participants is a critical task in many autonomous driving systems, particularly for constructing realistic simulation to test the AD planners. 
Most existing simulators~\citep{Dosovitskiy17,highway-env,zhou2020smarts} rely on hand-crafted rules for traffics and maps generation. However, the data distributions for the map structure, traffic flow, the interaction between traffic participants and other elements do not realistically represent the real world.
Modern data-driven simulators~\citep{vinitsky2022nocturne,gulino2023waymax,li2021metadrive,li2023scenarionet} address this by replaying the behaviors of the traffic participants from real-world scenarios recorded by an autonomous vehicle (log-replay).
Yet, the downside of log-replay is that re-simulation may become unrealistic when the planner behavior diverges from the original logged behavior. For example, if an AD planner is more cautious than the human driver and brakes earlier,
the trailing vehicle might collide into it, leading to a false positive collision.

In this work, we mainly focus on the simulation agents task and evaluate our solutions on the Waymo Open Sim Agent Challenge (WOSAC)~\citep{montali2023wosac}. Many existing WOSAC submissions apply the 
marginal motion prediction models~\citep{shi2022motion,feng2023trafficgen,bergamini2021simnet}, which typically take initial states and predict the positions of traffic participants at all future steps in a single inference (one-shot). Those marginal models do not explicitly model interactions between agents during the prediction horizon.
The autoregressive (AR) models naturally fit to the driving behavior modeling, especially in the context of closed-loop simulation~\citep{suo2021trafficsim,zhang2023trafficbots,seff2023motionlm,kamenev2022predictionnet}. 
AR decoding~\citep{seff2023motionlm} allows the interactions between agents to be modeled via a self-attention mechanism at each step of the decoding process.
However, closed-loop training of the AR behavior prediction models remains an understudied area.
We propose to improve a pre-trained AR model with closed-loop fine-tuning and evaluate the performance on the WOSAC benchmark.

In contrast to prior research on combining behavior cloning and reinforcement learning~\citep{lu2023imitation,zhang2022trajgen,zhang2023learning}, our approach eliminates the need for an external simulator and a dynamics model. Since our model predicts actions for all agents, it functions as a simplified simulation environment itself.
From an algorithmic perspective, we avoid back-propagation through time (BPTT) used in existing works~\citep{kamenev2022predictionnet,zhang2023learning}. Instead, we propose that RL can be conducted with a minimalistic policy gradient algorithm~\citep{williams1992simple}. This allows us to use non-differentiable rewards (such as a boolean collision indicator) and non-differentiable models with discrete outputs which would not be possible with BPTT.


\section{Preliminaries}

\paragraph{Motion Prediction.}

A driving scenario includes static information such as the map topology and dynamic information such as the states of traffic participants and traffic lights.
At each time step, the state of a traffic participant is represented by a feature vector containing the position, velocity, and heading angle in the global frame, and the object type (vehicle, cyclist, pedestrian). For traffic lights, the feature vector contains their position and state (green, yellow, red or unknown).
Given the history states of $N$ traffic participants with indices $\cI = [1 ... N]$, the goal of motion prediction is to predict future trajectories, i.e. the positions in future steps, of these agents.

\paragraph{Behavior Modeling as a Multi-Agent RL Problem.}
We consider driving behavior modeling as a Multi-agent Markov Decision Process (Multi-agent MDP).
The Multi-agent MDP is defined by the tuple $\langle \cI,$ $\cS,$ $\{\cA_i \},$ $\mathcal{T},$ $\{\mathcal{R}_i \},$ $\Omega,$ $\{\mathcal{O}_i\},$ $\gamma \rangle$,
where
$\mathcal S$ is the joint state space,
$\mathcal A = \times_{i} \mathcal{A}_i$ is the joint action space,
the transition function is $\mathcal T: \mathcal S \times \mathcal A \to \mathcal S$,
the reward functions are $\mathcal R_i(s_t, a_{t, i}, s_{t+1}), \forall i \in \mathcal I$,
the observation space is $\Omega = \times_{i} \Omega_i$, the observation functions 
$\mathcal O_i(s)$, and the discount factor is $\gamma$.
In this Multi-agent MDP, 
the goal is to learn action policies $\pi_i: \mathcal{A}_i \times {\Omega_i}\to[0,1]$ for each agent.
Each agent aims to maximize its expected cumulative return:
$
\pi_i = \arg\max_{\pi} \expect_{\tau\sim P_{\pi_j, \forall j \in \cI}}[
\sum_{t=1}^{T} \gamma^{t} r_{t, i}],
$
where $\tau = (\bs_0, \ba_0, ..., \bs_T, a_T)$ is the joint future rollout obtained by executing the learned policy model conditioned on the initial state.
Here, $\ba_t = \{ a_{t, i} \}_{i \in \cI}$ is the joint actions.

\paragraph{Autoregressive Encoder-Decoder Architecture.}

\begin{wrapfigure}{r}{0.35\textwidth}
\centering
\vspace{-1em}
\includegraphics[width=\linewidth]{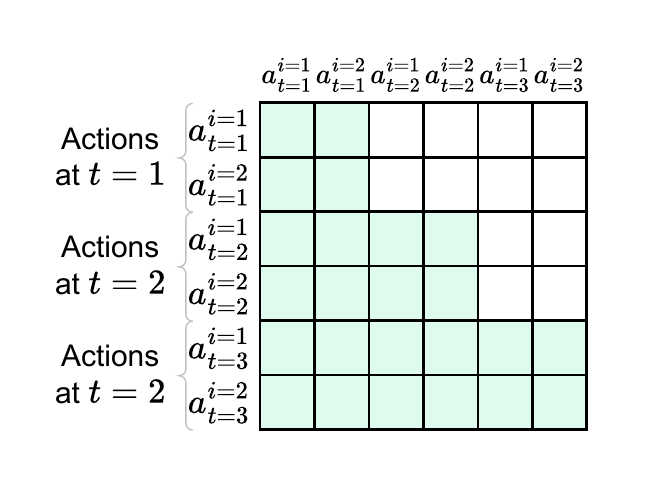}
\caption{The causal mask in the decoder.}
\label{fig:casual_mask}
\vspace{-1em}
\end{wrapfigure}

We use MotionLM~\citet{seff2023motionlm}, an encoder-decoder transformer-based autoregressive motion prediction model.
The model's encoder takes a set of tokens representing the initial states of the scenarios as input and generates a scene embedding. These initial states include the traffic lights states, map topology, and the history information of all traffic participants.
During inference, we run the decoder for $T_\text{pred}$ prediction steps to autoregressively generate the prediction of all agents.
At each prediction step, the decoder takes a set of motion tokens as well as the scene embedding as input and generates a distribution of $N$ output tokens.
The decoder consists of multiple layers, each applying self-attention among input tokens and cross-attention to the scene embedding. All $N$ motion tokens at step $t$ can attend to each other and all previous tokens as shown in Fig.~\ref{fig:casual_mask}, where each row represents a query token and each column a key token and green blocks indicate key tokens that the query can attend.
After running $T_\text{pred}$ prediction steps, the $T_\text{pred}\times N$ output motion tokens form complete trajectories for $N$ agents. 
This autoregressive approach ensures that each agent's action is based on a temporally causal relationship with the previous actions of all traffic participants, leading to improved modeling of interaction between agents within the prediction horizon.

We modify the original MotionLM model by adopting a scene-centric input format and
predicting the motion of all agents, rather than using a pre-selected subset.
The scene-centric representation and the encoder-decoder architecture enable more computationally efficient encoding of the scene context and prediction of all agents' motion.


\section{Method}
\label{sec:method}

As shown in Algorithm~\ref{algorithm}, our method has two stages.
In the pre-training stage, we reconstruct the ground truth actions from the data and use the maximum likelihood objective to match the joint action distribution of observed behaviors in the dataset:
\begin{equation}
\label{eq:pre-training-loss}
\max_{\pi_{\theta}} \expect_{\mathcal D} \sum_{t=1}^{\Tpred} \sum_{i \in \cI} \log\pi_\theta(a^{\text{GT}}_{t, i} | o_{t, i}).
\end{equation}


\noindent The second stage of our learning process fine-tunes the model using reinforcement learning (RL).
We formalize the problem as a Multi-agent MDP for our behavior modeling task as follows:

\noindent
\textbf{Action.}
The action space for each agent $\mathcal{A}_i$ is a Verlet-wrapped delta action space~\cite{seff2023motionlm}, where each action represents the X, Y acceleration in scene coordinates. To reconstruct the ground truth action targets, we first infer the accelerations by differentiating the observed positions in the data. These accelerations are then discretized into a 13x13 uniformly spaced grid, where outliers are clipped to the minimum and maximum values of 6 $m/s^2$.

\noindent \textbf{State.}
The state space $\mathcal S$ contains map features, the joint state of all objects and traffic lights.

\noindent \textbf{Transition Dynamics.}
Agents transit to new positions computed by adding previous positions with an offset: $pos_{t+1, i} = (a_{t, i} \Delta + vel_{t, i})\Delta + pos_{t, i}$ wherein the velocity $vel_{t, i}$ and the position $pos_{t, i}$ are in $s_{t}$ and $\Delta$ is the time interval between steps.

\noindent \textbf{Observation.}
We define observations to consist of a historic context $c$, previous actions of all objects and the agent identity: $o_{t,i} = (c, \mathbf a_1, ..., \mathbf a_{t-1}, i)$.
Here, the context  $c = (\{m_i\}_M, s_{T_\text{prev}}, ..., s_0)$ is a set containing $M$ map features and the object and traffic light states for history steps $t= T_{\text{prev}}, ..., 0$.

\subsection{RL Fine-tuning}

We propose to fine-tune a pre-trained autoregressive motion predictor with RL. The reward function for each agent at each step is defined as:
\begin{equation}
\label{eq:reward-function}
    r_{t, i} = 
    -|| Pos_{t, i} - GT_{t, i} ||_2 - \lambda Coll_{t, i}
    ,
\end{equation}
where $Pos_{t, i}$ is the position of agent $i$ in step $t$ and $GT_{t, i}$ is the corresponding position in the logged trajectory. $Coll_{t, i}$ is a Boolean indicator and will be 1 if the bounding box of agent $i$ intersects with others' bounding boxes.
This reward function, while simple, captures the key objectives of preserving the behavioral realism as well as satisfying the safety constraint of collision avoidance.

\RestyleAlgo{ruled}
\begin{algorithm}[!t]
\SetKwInOut{Input}{input}
\SetKwInOut{Output}{output}
\caption{Pre-train and fine-tune an autoregressive motion predictor.}
\label{algorithm}
\SetAlgoLined
\LinesNumbered
\DontPrintSemicolon
\Input{Large-scale driving dataset $\mathcal D$.}
\Output{A Sim Agent policy $\pi_\theta$.}
Initialize model parameters $\theta$ and model $\pi_\theta$. \\
\For{pre-training iterations $j=1, ...$}{
Retrieve $\bshistory, \bsGT$ from $\mathcal D$. \\
Construct target actions $\ba_{\text{GT}} = \{a^\text{GT}_{t, i}, \forall t, i\}$ and construct the observation $o_{t, i}$ with GT actions. \\
Run the model with the observation and get the predicted actions $\{a_{t, i}, \forall t, i\}$. \\
Update $\pi_\theta$ via Eq.~\ref{eq:pre-training-loss}. \\
}
\For
{fine-tuning iterations $j=1, ...$}
{
Retrieve $\bshistory$ from $\mathcal D$; Set $\ba_0 \gets \o$. \\
\For
(\Comment*[f]{Autoregressive Rollout.})
{$t = 1, ..., \Tpred$}
{
$\bo_{t} \gets (\bshistory, \ba_{0}, ..., \ba_{t-1}, \cI)$. \Comment*{Get obs.}
$\ba_{t} \gets \pi_\theta(\cdot | \bo_t)$. \Comment*{Decode next actions.}
}
Reconstruct predicted states $\bspredict = \{ \hat{s}_{t, i}, \forall t, i \}$ by translating actions $a_{t, i}, \forall t, i$. \\
Compute per-agent per-step $r_{t, i}$ via Eq.~\ref{eq:reward-function} \\
Compute normalized return via Eq.~\ref{eq:normalized-return} and update $\pi_\theta$ via Eq.~\ref{eq:policy-gradient}
}
\end{algorithm}

During fine-tuning, we run the model for $\Tpred$ prediction steps.
The encoder first encodes the scene context $c$ as a shared scene embedding, before the autoregressive decoding.
At each prediction step $t$, the fixed scene embedding and the $t \times N$ tokens are fed to the autoregressive decoder and sample $N$ new actions.
Specifically, at prediction step $t=1$, we project the agents' current positions through a MLP and get the agent embeddings:
$id_i = \text{MLP}(Pos_{0, i}), i=1, ..., N$. The agent and scene embeddings serve as the input tokens to the decoder.
After several layers of self-attention and cross-attention, 
$N$ actions are sampled from the categorical distributions constructed from the output of the decoder. 
The embeddings of those sampled actions will be added with corresponding $id_i$ and concatenated with the tokens in previous steps to form the input tokens for the next step.
Compared to the decoding process of a language model, we output $N$ tokens concurrently at each prediction step instead of one token.
Our model autoregressively rolls out the actions in $\Tpred$ time steps.
After collecting the rollout trajectories, we translate the actions to sequences of 2D positions for computing the rewards following Eq.~\ref{eq:reward-function}.
The return (\textit{i.e.} the ``reward-to-go''), for step $t$ and each agent $i$ is:
\begin{equation}
R_{t, i} = \sum_{t' = t}^{\Tpred} \gamma^{t'-t} r_{t', i}.
\end{equation}
We normalize the return across the training batch, here $Mean$ and $Std$ are the average and the standard deviation computed across all time steps in all scenarios for all agents in the training batch:
\begin{equation}
\label{eq:normalized-return}
\tilde{R}_{t, i}  = (
{R}_{t, i} - Mean(R)
) / Std({R}).
\end{equation}
We then apply the REINFORCE~\citep{williams1992simple} method to compute a policy gradient for optimizing the model by differentiating the following surrogate objective:

\begin{equation}
\label{eq:policy-gradient}
\max_{\pi_{\theta}} \expect_{\mathcal D} \sum_{t=1}^{\Tpred} \sum_{i \in \cI} \log \pi_{\theta}( a_{t, i} | o_{t, i} )  \tilde{R}_{t, i}.
\end{equation}

\subsection{Policy Evaluation for Sim Agents}
\label{sec:policy_eval}

A key limitation of common ``imitative'' metrics (such as ADE), which compare model rollouts to ground truth trajectories, is the weak connection between the metric and the actual goal of assessing the AD planner performance. A low ADE metric does not guarantee good driving behaviors. Log-replay, for example, has a perfect ADE of zero but would be a poor choice for sim agents because it is non-reactive. 
To create an evaluation that has a direct connection to measuring the performance of the AD planners, we propose a new policy evaluation framework for sim agents, inspired by the RL policy evaluation literature ~\citep{uehara2022ope}.

\begin{figure}[!t]
\includegraphics[width=\linewidth]{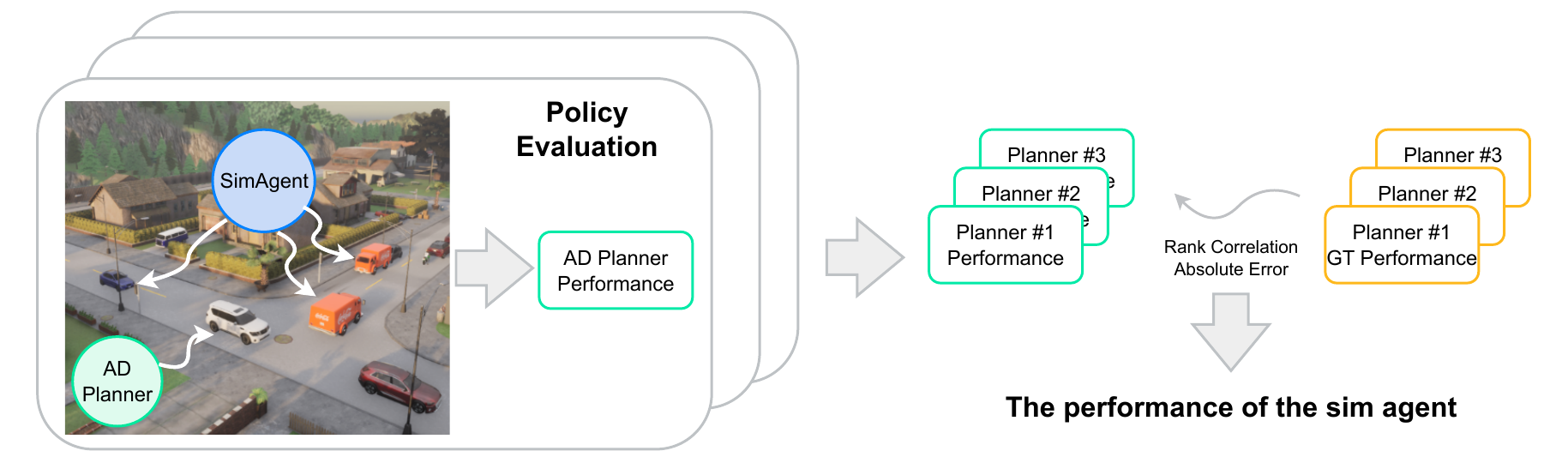}
\caption{
We evaluate the sim agent by its ability to correctly assess the AD planner.
}
\label{fig:policy-eval}
\end{figure}

Our policy evaluation framework involves ranking and scoring the performance of a predetermined collection of AD planner policies. This is analogous to a real-world use case where one must decide which planner to deploy from a collection of candidate software releases. A better sim-agent model will give a more accurate signal on which policy would be best when deployed in the real world. As shown in Fig.~\ref{fig:policy-eval}, we first prepare a batch of AD planner policies with known performance ranking. Then, we evaluate the performance of these AD planners when the traffic agents in the scenario are controlled by a sim agent model. Therefore, we will generate the estimated performance for those AD planners for the specific sim agent. We then measure the discrepancy between the estimated performance and the ground truth performance of those planners. This discrepancy becomes the measurement of the sim agents model's ability to assess the performance of the planners. 
Policy evaluation covers two important use cases for the sim agent models in the deployment of autonomous vehicles: \textit{evaluation}, where we wish to estimate the performance of agents in the simulation, and \textit{selection}, where we wish to determine a ranking or order between different deployment candidates.

\paragraph{Choice of policies.} In order to perform policy evaluation, we must have a fixed set of policies on hand to rank or evaluate. To generate a large variety of planning policies with both good and bad performance, we propose to use a random shooting search-based policy family, parameterized by the number ($J$) and depth ($D$) of trajectories sampled. We compute a ``ground truth'' score for each policy by evaluating it with log playback agents. Note that the choice of ground truth is an important design decision. Any sim agent could serve as a ground truth, but we need to pick one that is the most fair to all models and we believe log playback to be the most neutral.

Our random shooting policy operates in a model-predictive control (MPC) fashion: at each time step, our random shooting policy samples from a fixed library of $J$ trajectories, which are generated by maintaining a single steering wheel angle and acceleration for $D$ steps. Note that this action specification is different from the action space of the MotionLM architecture we described in Sec.~\ref{sec:method}. We found this simple strategy to work much better than randomly selected actions.
The trajectories are then scored by a reward function and the first step of the best scoring trajectory is executed. This process repeats for the entirety of the rollout. We used 16 different settings of $J$, ranging between 9 to 81, and we used 4 values of depth $D \in [6, 8, 12, 16]$. We then used the product of these two sets, for a total of 64 different policies evaluated.

\paragraph{Reward Function.} The reward function used for selecting actions from a set of candidate trajectories is a linear combination of collisions, as well as off-road and route-following infractions. We use a modified reward function from Eq.~\ref{eq:reward-function} by replacing the L2 norm from the ground truth (which is not available to the planner at execution time) with additional terms for following a reasonable path. We instead give the planner a high-level route in the form of waypoints, and we use a weighted sum of $-10C - O - R + 10^{-4}P$, where 
$C\in \{0, 1\}$ denotes the collision indicator and is 1 when a collision between AV and another object happens,
$O\in \{0, 1\}$ denotes the offroad indicator and is 1 when the AV is too close to the road edge, 
$R \in \{0, 1\}$ denotes the off-route indicator which is 1 when the AV's lateral distance to the GT trajectory exceeds a threshold, and $P$ is the projection of the AV's displacement between two time steps when projected onto the logged trajectory and measures the route-following behavior.
We use Waymax~\cite{gulino2023waymax}'s utility function to compute those metrics.

\section{Experiments}

\begin{table*}[t]
\footnotesize
\setlength{\tabcolsep}{1mm}
\centering
\resizebox{\textwidth}{!}{%
\begin{tabular}{c|ccccccccc|ccc}
\specialrule{.2em}{.1em}{.1em}
 &
\shortstack{Lin. Speed\\$\uparrow$ } & 
\shortstack{Lin. Acc.\\$\uparrow$ } & 
\shortstack{Ang. Speed\\$\uparrow$ } & 
\shortstack{Ang. Acc.\\$\uparrow$ } & 
\shortstack{Dist. to\\Obj. $\uparrow$} & 
\shortstack{Collision \\$\uparrow$ } & 
\shortstack{TTC \\ $\uparrow$ } & 
\shortstack{Dist. to \\ Road Edge $\uparrow$ }& 
\shortstack{Offroad \\$\uparrow$ }& 
\shortstack{Composite \\ $\uparrow$ }& 
\shortstack{ADE \\ $\downarrow$}& 
\shortstack{MinADE \\$\downarrow$} 
\\
\hline 
Random
& 0.002 & 0.044 & 0.074 & 0.120 & 0.000 & 0.000 & 0.734 & 0.178 & 0.287 & 0.155 & 50.739 & 50.706 \\
Constant Velocity
& 0.074 & 0.058 & 0.019 & 0.035 & 0.208 & 0.345 & 0.737 & 0.454 & 0.455 & 0.287 & 7.923 & 7.923 \\
Wayformer
& 0.331 & 0.098 & 0.413 & 0.406 & 0.297 & 0.870 & 0.782 & 0.592 & 0.866 & 0.575 & \textbf{2.498} & 2.498 \\
MVTE
& 0.445 & 0.222 & 0.535 & 0.481 & 0.383 & 0.893 & 0.832 & 0.664 & 0.908 & \textbf{0.645} & 3.859 & \textbf{1.674}\\
Logged Oracle
& 0.561 & 0.330 & 0.563 & 0.489 & 0.485 & 1.000 & 0.881 & 0.713 & 1.000 & 0.722 & 0.000 & 0.000\\
\hline 
Pre-trained (1M) & 0.390 & 0.235 & 0.504 & 0.447 & 0.348 & 0.544 & 0.803 & 0.582 & 0.525 & 0.490 & 6.332 & 3.177 \\
RL-only (1M) & 0.257 & 0.115 & 0.487 & 0.429 & 0.244 & 0.239 & 0.759 & 0.456 & 0.164 & 0.320 & 7.785 & 6.918 \\
Fine-tuned (1M) & 0.412 & 0.219 & 0.451 & 0.420 & 0.348 & 0.863 & 0.814 & 0.637 & 0.804 & 0.597 & 2.436 & 1.867 \\
Pre-trained (10M) & 0.439 & 0.241 & 0.502 & 0.454 & 0.371 & 0.673 & 0.811 & 0.625 & 0.655 & 0.549 & 4.508 & 2.274 \\
Fine-tuned (10M) & 0.433 & 0.220 & 0.455 & 0.423 & 0.361 & 0.877 & 0.819 & 0.647 & 0.825 & \textbf{0.608} & \textbf{2.428} & \textbf{1.706} \\
\hline
\end{tabular}
}
\caption{\label{tbl:wosac_results}Results on the Waymo Open Sim Agents Challenge (WOSAC) benchmark. Metrics marked with ($\uparrow$) are better if higher, while metrics marked with ($\downarrow$) are better if lower. Fine-tuned agents score better than pre-trained agents on safety-critical metrics such as collision and offroad, which results in a significantly higher composite metric score. We bold the best results for the baseline agents and best results from the model we trained for this work.
}
\end{table*}

We now describe our method's experimental results on the Waymo Open Sim Agents Challenge (WOSAC)~\cite{montali2023wosac} and on the Policy Evaluation task introduced in Sec.~\ref{sec:policy_eval}. Our experiments are designed to answer the following questions:
\begin{enumerate}
    \item Does RL fine-tuning improve the overall sim agents behavior?
    \item Can fine-tuning be used to improve targeted metrics through reward engineering?
    \item Does the fine-tuned sim agents model provide better evaluation of AD planner performance?
\end{enumerate}

\paragraph{Dataset.} We train our method and baselines on the Waymo Open Motion Dataset (WOMD)~\citep{waymo_open_dataset} and evaluate on the Waymo Open Sim Agents Challenge (WOSAC) benchmark~\citep{montali2023wosac}. WOMD contains scenarios recorded at 10Hz including one second of history (11 discrete time steps), and 8 seconds of future states (80 time steps) to  predict. In total, there are 486k training scenarios, 44k validation scenarios, and 45k test scenarios. Up to 128 agents are simulated in each scenario.

\paragraph{Model.}
We use a pre-trained autoregressive motion prediction model MotionLM~\cite{seff2023motionlm}.
For the MotionLM model with 10M parameters, we use 4 encoder and 4 decoder layers. The hidden size is 256. The number of attention heads is 4. The activation is ReLU. The feed-forward network intermediate size is 1024.

\paragraph{Training.}
We pre-train a MotionLM model on the WOMD training set with the objective in Eq.~\ref{eq:pre-training-loss}. The model is then used for RL fine-tuning. 
The encoder and the decoder of the model are fine-tuned jointly.
1M steps of updates are conducted both in pre-training and fine-tuning. At each training step, 128 scenarios are sampled from the dataset to form a batch.
During RL fine-tuning, the learning rate is set to 5e-6 and the discount factor is set to 0.95.

\subsection{Waymo Open Sim Agents Challenge}

\paragraph{Baselines.} We include several notable baselines reported by~\citet{montali2023wosac} on WOSAC. The ``random'' and ``constant velocity'' agents are included to provide a reasonable performance lower bound. The ``logged oracle'' represents the ground truth future behavior that is not visible to other baselines and represents an upper-bound on performance. Wayformer~\citep{nayakanti2023wayformer} is a recent transformer-based model which shares the same encoder structure as our model. MVTE~\citep{wang2023multiverse} is another transformer-based architecture which is the current state-of-the-art on the benchmark. 
Both Wayformer and MVTE adopt the agent-centric input representation.
We also report the performance of our pre-trained 1M parameter and 10M parameter models, which are based on the MotionLM~\citep{seff2023motionlm} architecture.

\paragraph{Evaluation Metrics.} The Waymo Open Sim Agents challenge evaluates agents on a wide range of \textit{likelihood}-based metrics. These metrics are designed to measure realistic simulation in aggregate over the entire dataset, while allowing agents to have enough flexibility to deviate from the exact logged ground truth in each scenario. Each benchmarked method samples 32 rollouts for each WOMD test scenario. Metrics (such as velocity and heading angle) are then measured on these samples, but binned into discrete histograms, and the log-likelihood of the ground-truth data is measured under these histograms. The individual scores are then weighted and averaged to produce a final composite metric. In addition, following WOSAC~\citep{montali2023waymo} we also report the mean average displacement error (ADE) over 32 rollouts and the minimum average displacement error (minADE) over 32 rollouts.

\begin{figure*}[tb]
    \centering
    \includegraphics[width=\textwidth]{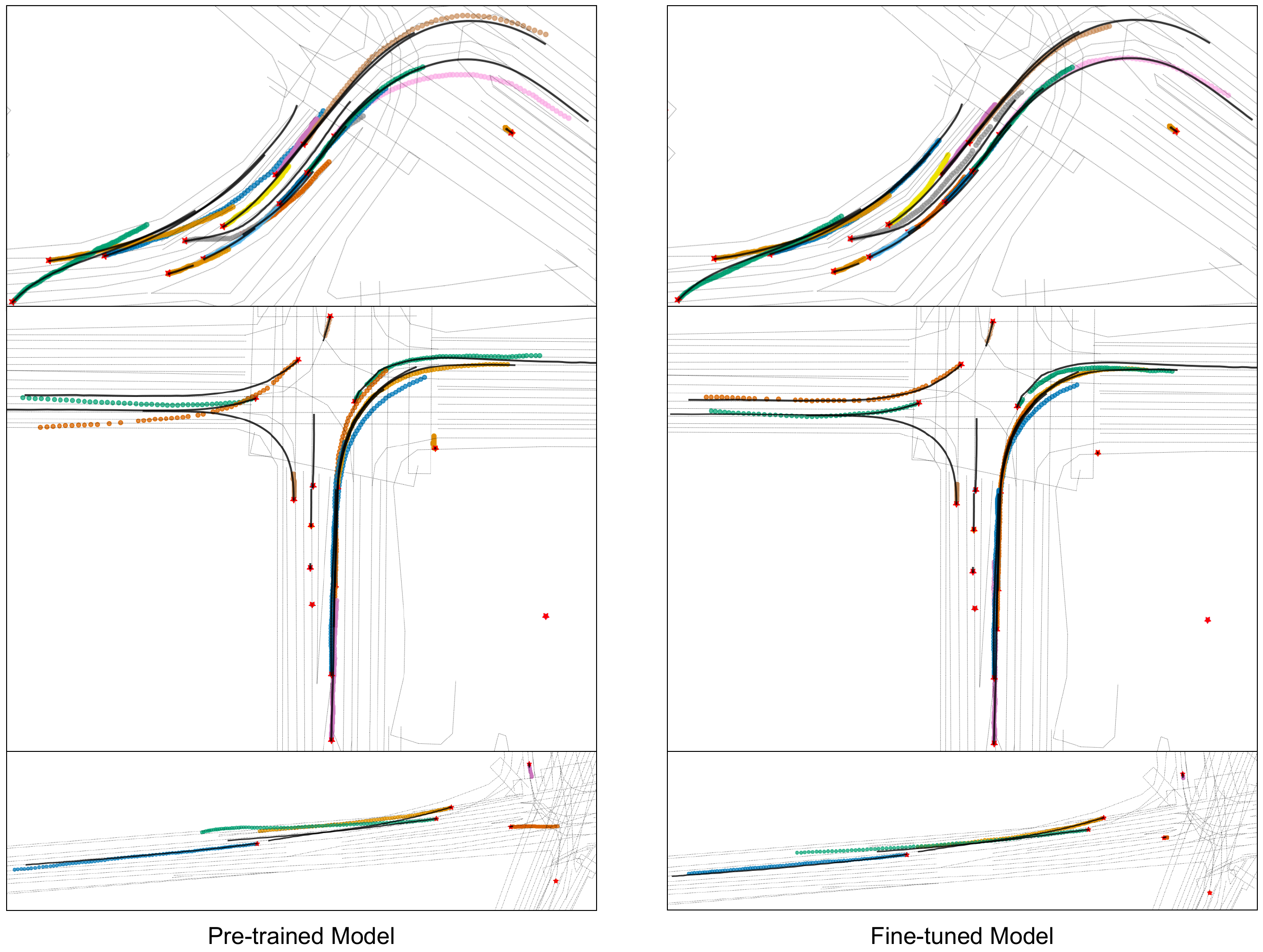}
    \caption{Visualization of scenario rollouts using a pre-trained and a fine-tuned model. The start locations of vehicles are marked with a red star, the ground truth futures are marked with a solid black line, and the sampled trajectory is marked with circles of different colors. \textbf{Left}: The pre-trained model suffers from drifting due to distributional shift between training (with teacher forcing) and testing (with an autoregressive rollout). \textbf{Right}: The fine-tuned model is able to follow the ground truth much more precisely, which is quantitatively demonstrated by the better ADE metric.}
    \label{fig:qualitative}
\end{figure*}

\paragraph{Results.} We report our results on WOSAC in Table~\ref{tbl:wosac_results}. The results provide strong evidence that closed-loop fine-tuning from a pre-trained model can significantly improve performance of the model as a sim agent. We see that both the ``Fine-tuned 1M'' and ``Finedtuned 10M'' models perform better than ``Pre-trained 1M'' and ``Pre-trained 10M'', respectively. When we look at the breakdown of the constituent metrics, we see that most of the gains come from improved safety-critical metrics such as collision and offroad. As a concrete example, Fig.~\ref{fig:qualitative} illustrates a single scenario comparing a rollout from a pre-trained 1M parameter model with a fine-tuned model. The pre-trained model is prone to slowly drifting away from the ground truth trajectory, a distribution shift problem commonly impacting pure imitative and teacher forcing~\citep{ross2011reduction} methods. By training the model in closed-loop, we can mitigate this distribution shift issue.

The composite performance of ``Fine-tuned 10M'' is still lower than the state-of-the-art MVTE~\citep{wang2023multiverse} model. We believe this can be mostly attributed to the choice of our pre-trained baseline model, which was our re-implemented version of the MotionLM~\citep{seff2023motionlm} architecture with the encoder architecture of Wayformer~\citep{nayakanti2023wayformer}.
We leave the fine-tuning of MVTE using the same methodology described in this work as future work, which we hypothesize would lead to a sizeable performance gain.

\begin{table}[!t]
\setlength{\tabcolsep}{1mm}
\centering
\begin{tabular}{c|cc|ccc}
\specialrule{.2em}{.1em}{.1em}
Collision Weight & Collision & TTC & Composite & ADE & MinADE \\
& $\uparrow$ & $\uparrow$ & $\uparrow$ & $\downarrow$ & $\downarrow$ \\
\hline 
0 & 0.834 & 0.810 & 0.590 & \textbf{2.405} & 1.871 \\
2 & \textbf{0.863} & 0.814 & \textbf{0.597} & 2.436 & \textbf{1.867} \\ 
5 & 0.844 & \textbf{0.817} & 0.595 & 2.838 & 1.980 \\ 
10 & 0.831 & \textbf{0.817} & 0.594 & 3.058 & 2.023 \\ 
\hline
\end{tabular}
\caption{\label{tbl:collision_ablation}WOSAC benchmark scores for different values of the collision fine-tuning weight $\lambda$ with the 1M parameter model. Increasing the collision weight improves the collision score at the cost of decreasing imitative behavior metrics such as ADE. We find a good balance at an intermediate value.
}
\end{table}

To measure the effect of reward engineering and the ability to target specific metrics with fine-tuning, we ran an ablation study varying the relative weight $\lambda$ of the collision metric in Eq.~\ref{eq:reward-function}, with results reported in Table~\ref{tbl:collision_ablation}. We fine-tuned the 1M parameter model with 4 different collision weights, and reported the collision score versus the ADE of the predictions. We can clearly see that adding some amount of collision penalty  improves the collision metric at the cost of degrading the ADE metric. This generally makes sense, as the optimization must trade-off the collision penalty versus the displacement error terms in the reward function. However, we also see that at very high values of the collision weight, all metrics tend to degrade, and the best result is with an intermediate cost weight. We hypothesize this is the case because displacement error is a very dense and rich reward signal, whereas collision is a more sparse and noisy signal.

\subsection{Policy Evaluation}
\label{sec:experiment-policy-eval}

As described in Sec.~\ref{sec:policy_eval}, we also propose to evaluate sim agents through the lens of policy evaluation.
In particular, we follow the methodology proposed by~\citet{fu2020benchmarks} and report two metrics used in their benchmark: Spearman's rank correlation and absolute error. Rank correlation is a metric for ``selection'' and measures the ability for the simulation to discern between good and bad policies (AD planners). This is useful for the problem setting where one must select the best policy to deploy from a set of candidates. On the other hand, absolute error is a metric for ``evaluation'' and measures how closely a simulation comes close to estimating the true cost or reward accrued by the policy. This is useful if one is interested in concrete performance numbers such as estimating the rate of a particular event of interest. The sim agent causing less absolute error and higher rank correlation is better for evaluating different planners.

Given a set of planners with known performance ranking, we assess sim agents by measuring how well they estimate the value function of each planner. We generate a Monte-Carlo estimate of each planner's value by running it on each scenario, where all traffic participants are controlled by the sim agent, and computing the empirical returns of the planner. The returns are then averaged across all scenarios in the WOMD test set to form the estimate of the planner. 
The value estimates are then compared to a ``ground-truth value'', and we show two quantities, \textit{rank correlation} and \textit{absolute error} in Table~\ref{tbl:policy_eval_rank_corr} and Table~\ref{tbl:policy_eval_abs_err}, respectively.
Each row in these tables represents the policy evaluation results of a sim agent. A sim agent will generate $K=64$ estimated returns of $K$ policies (AD planners). Due to the absence of real-world simulation, there is no a universal ground truth sim agent that can fully replicate the realistic behaviors. We can not know the ground truth returns when running these $K$ policies in the real world. In this work, we picked the log-playback sim agent as the ``ground truth sim agent'' and considered the average returns of these $K$ policies when running with the log-playback sim agent as the ground truth value. These form the \texttt{Log} column in Table~\ref{tbl:policy_eval_rank_corr} and Table~\ref{tbl:policy_eval_abs_err}. 
We also show rank correlation and absolute error relative to each other sim agent we considered for completeness.


\begin{table}[!t]
\setlength{\tabcolsep}{1mm}
\centering
\begin{tabular}{c||c|cccc}
\specialrule{.2em}{.1em}{.1em}
Sim Agent & Log & Pre. 1M & Fine. 1M & Pre. 10M & Fine. 10M\\
\hline
Pre-trained 1M & 0.859 & - & 0.954 & 0.953 & 0.911 \\
Fine-tuned 1M & 0.865 & 0.954 & - & 0.959 & 0.925 \\
Pre-trained 10M & 0.845 & 0.953 & 0.959 & - & 0.932 \\
Fine-tuned 10M & \textbf{0.866} & 0.911 & 0.925 & 0.932 & - \\
\hline
\end{tabular}
\caption{\label{tbl:policy_eval_rank_corr}Policy evaluation \textbf{rank correlation} results for pre-trained and fine-tuned models (higher is better). Each cell corresponds to a ranking correlation of estimated returns of a set of predefined AD planners between two sim agent models. For example, the highlighted \texttt{Fine-tuned 10M - Log} means the ranking correlation of predefined AD planners when using the Fine-tuned 10M and the log-playback sim agents.
}
\end{table}

\begin{table}[!t]
\setlength{\tabcolsep}{1mm}
\centering
\begin{tabular}{c||c|cccc}
\specialrule{.2em}{.1em}{.1em}
Sim Agent & Log & Pre. 1M & Fine. 1M & Pre. 10M & Fine. 10M\\
\hline
Pre-trained 1M & 10.518 & - & 0.557 & 0.551 & 1.013 \\
Fine-tuned 1M & 10.101 & 0.557 & - & 0.426 & 0.729 \\
Pre-trained 10M & 10.014 & 0.551 & 0.426 & - & 0.602 \\
Fine-tuned 10M & \textbf{9.509} & 1.013 & 0.729 & 0.602 & - \\
\hline
\end{tabular}
\caption{\label{tbl:policy_eval_abs_err}Policy evaluation \textbf{rank absolute error} results for pre-trained and fine-tuned models (lower is better). 
}
\end{table}

\paragraph{Results.}
Our results are reported in Table~\ref{tbl:policy_eval_rank_corr} and Table~\ref{tbl:policy_eval_abs_err}. We report results on a total of 64 candidate policies created by varying the depth (D) and the number of sampled trajectories (J).
According to the \texttt{Log} column, we see that with fine-tuned models, a higher rank correlation and lower absolute error relative to the log-playback sim agent can be achieved, indicating that the fine-tuned models are more accurate at measuring a planner's performance and deciding whether the planner is better than another.

\section{Conclusion}
We studied the viability of applying the popular ``pre-training and fine-tuning'' scheme to modeling traffic agents for AD simulation.
We drew the connection between a multi-agent driving behavior model and a simulation environment -- the multi-agent behavior model itself can be used to perform rollouts for closed-loop training.
By using an on-policy RL algorithm with a simple reward, we are able to fine-tune a pre-trained large multi-agent behavior model to effectively align the traffic agent behaviors with human expectations, such as collision avoidance.
The experimental results show that our method can significantly improve the performance of the pre-trained model on the Waymo Open Sim Agent Challenge (WOSAC)~\citep{montali2023wosac}. 
We also proposed a novel policy evaluation task and demonstrated that the model fine-tuned by our method can achieve more reliable AD testing result.

\noindent \textbf{Limitations.} 
There are several limitations to the approach we have discussed in this paper. We use a simple transition and action model (based on predicting accelerations and integrating them to estimate positions) as the environment dynamics model, which could produce kinematically unrealistic behaviors during a rollout. A more realistic solution would be to embed a low-level controller into simulation that attempts to reach the positions predicted by the model.
In addition, we studied a reward function (Eq.~\ref{eq:reward-function}) that encourages collision avoidance and minimizes divergence in closed-loop simulation. The reward function can be extended to induce various driving behaviors, such as encouraging adversarial behavior (e.g. using the negative of the ego vehicle's reward) to stress-test challenging scenarios.
\bibliographystyle{splncs04}
\bibliography{main}



\end{document}